\documentclass{article}

\usepackage{microtype}
\usepackage{graphicx}
\usepackage{subfigure}
\usepackage{booktabs}
\usepackage{bbm}
\usepackage{subfigure}
\usepackage{multirow}
\usepackage{color,verbatim}
\usepackage[tableposition=top]{caption}
\usepackage{hyperref}
\usepackage{enumitem}
\usepackage{amssymb}
\usepackage{amsmath}
\usepackage{amsthm}
\usepackage{blindtext}
\usepackage{xcolor}

\newtheorem{theorem}{Theorem}[section]
\newtheorem{corollary}{Corollary}[section]
\theoremstyle{definition}

\usepackage{array}
\newcolumntype{L}[1]{>{\raggedright\let\newline\\\arraybackslash\hspace{0pt}}m{#1}}
\newcolumntype{C}[1]{>{\centering\let\newline\\\arraybackslash\hspace{0pt}}m{#1}}
\newcolumntype{R}[1]{>{\raggedleft\let\newline\\\arraybackslash\hspace{0pt}}m{#1}}

\usepackage{hyperref}

\usepackage[accepted]{icml2020}

\icmltitlerunning{Few-shot Relation Extraction via Bayesian Meta-learning on Relation Graphs}
\begin{document}

\twocolumn[
\icmltitle{Few-shot Relation Extraction via Bayesian Meta-learning on Relation Graphs}




\begin{icmlauthorlist}
\icmlauthor{Meng Qu}{mila,udem}
\icmlauthor{Tianyu Gao}{thu}
\icmlauthor{Louis-Pascal A.~C.~Xhonneux}{mila,udem}
\icmlauthor{Jian Tang}{mila,hec,cifar}
\end{icmlauthorlist}

\icmlaffiliation{mila}{Mila - Quebec AI Institute, Montr\'eal, Canada}
\icmlaffiliation{udem}{University of Montr\'eal, Montr\'eal, Canada}
\icmlaffiliation{hec}{HEC Montr\'eal, Montr\'eal, Canada}
\icmlaffiliation{cifar}{CIFAR AI Research Chair}
\icmlaffiliation{thu}{Tsinghua University, Beijing, China}

\icmlcorrespondingauthor{Meng Qu}{meng.qu@umontreal.ca}
\icmlcorrespondingauthor{Jian Tang}{jian.tang@hec.ca}
%\icmlcorrespondingauthor{Eee Pppp}{ep@eden.co.uk}

\icmlkeywords{Machine Learning, ICML}

\vskip 0.3in
]



\printAffiliationsAndNotice{\icmlEqualContribution} % otherwise use the standard text.

\begin{abstract}

This paper studies few-shot relation extraction, which aims at predicting the relation for a pair of entities in a sentence by training with a few labeled examples in each relation. To more effectively generalize to new relations, in this paper we study the relationships between different relations and propose to leverage a global relation graph. We propose a novel Bayesian meta-learning approach to effectively learn the posterior distribution of the prototype vectors of relations, where the initial prior of the prototype vectors is parameterized with a graph neural network on the global relation graph. Moreover, to effectively optimize the posterior distribution of the prototype vectors, we propose to use the stochastic gradient Langevin dynamics, which is related to the MAML algorithm but is able to handle the uncertainty of the prototype vectors. The whole framework can be effectively and efficiently optimized in an end-to-end fashion. Experiments on two benchmark datasets prove the effectiveness of our proposed approach against competitive baselines in both the few-shot and zero-shot settings. 

\end{abstract}

\section{Introduction}

A fundamental problem in natural language processing is relation extraction, which aims to identify the relations between entities in sentences. The problem is usually studied as a supervised classification task by training with labeled sentences. However, annotating a large set of sentences is time-consuming and expensive. As a result, the number of labeled data is very limited for this task. In practice, a common solution to this challenge is distant supervision~\cite{mintz2009distant}, where a knowledge graph is utilized to automatically generate training data. For example, given a triplet (\emph{Washington}, \emph{Capital}, \emph{US}) in a knowledge graph, all the sentences containing the two entities \emph{Washington} and \emph{US} will be labeled with the relation \emph{Capital}. In this way, a large quantity of training data can be generated for each relation, and such an approach has been extensively studied and has been proven very effective. However, a limitation of distant supervision is that the generated training data can be very noisy. This is because there could be multiple relations between two entities, and it is hard to determine which relation the entity pair belongs to in a particular context, or whether there is any relation expressed by the sentence.

An alternative approach for relation extraction, which is attracting growing interest, is meta-learning for relation extraction~\cite{han2018fewrel,gao2019fewrel}. The idea of meta-learning is to train models with a large number of diverse tasks, each of which has a few examples for demonstration, so that the learned model can quickly generalize to new tasks with only a few examples. For example, the Model-Agnostic Meta-Learning (MAML) algorithm~\cite{finn2017model} tries to find a good initialization for the parameters of a neural model, based on which the model can be quickly adapted to a new task with several steps of gradient descent. Another example is the prototypical network~\cite{snell2017prototypical}, which learns a prototype vector from a few examples for each class, and further uses the prototype vectors for prediction. Based on these techniques, handful recent studies~\cite{han2018fewrel,gao2019fewrel} are able to train relation extraction models with only a few examples for each relation. Although these methods achieve encouraging improvements, the performance remains unsatisfactory as the amount of information in training data is still limited. 

To more effectively generalize to new relations and tasks, in this paper we study modeling the relationships between different relations, and propose to leverage a global graph between different relations. In practice, such a global graph can be obtained in different ways. For example, we can use the knowledge graph embedding algorithms~\cite{bordes2013translating,sun2019rotate} to infer the relation embeddings and then construct a K-nearest neighbor graph based on the relation embeddings. The global relation graph provides the prior knowledge on the relationships between different relations, which allows us to transfer supervision between these relations and even generalize to these relations without leveraging any labeled sentences (i.e. zero-shot learning). 

Moreover, we propose a novel Bayesian meta-learning approach for few-shot relation extraction, which solves the problem by learning the prototype vectors of relations based on the labeled sentences (a.k.a.\ support set) and the global relation graph. Instead of learning a point estimation as in MAML~\cite{finn2017model} or prototypical networks~\cite{snell2017prototypical}, we follow existing work on Bayesian meta-learning~\cite{gordon2018meta,kim2018bayesian} and aim to infer the posterior distribution of the prototype vectors, which is able to effectively handle the uncertainty of the vectors. The posterior can be naturally factorized into a likelihood function on the support set, and a prior of prototype vectors obtained from the global graph. We parameterize the prior distribution of prototype vectors of relations by applying a graph neural network~\cite{kipf2016semi} to the global graph, allowing us to effectively utilize the relationships between different relations encoded in the graph.

For the posterior distribution of prototype vectors, existing studies~\cite{gordon2018meta,ravi2018amortized} usually parameterize it as a Gaussian distribution, and amortized variational inference is generally used to learn the posterior distribution. However, the posterior distribution of prototype vectors can be much more complicated than the Gaussian distribution. In this paper, instead of assuming a specific distribution for the posterior distribution of prototype vectors, we propose to directly sample from the posterior distribution with the stochastic gradient Langevin dynamics technique~\cite{welling2011bayesian}, which is very general and can be applied to different distributions. Our approach can be viewed as a stochastic version of MAML~\cite{finn2017model}, where random noises are added at each step of the gradient descent to model the uncertainty of prototype vectors. The sampled prototype vectors are further used to make predictions for query sentences in test sets, and the whole framework can be optimized in an end-to-end fashion.

We conduct extensive experiments to evaluate the proposed approach on two benchmark datasets of few-shot relation extraction. Empirical results prove the effectiveness of our proposed approach over many competitive baselines in both the settings of few-shot and zero-shot relation extraction.

\section{Related Work}

\subsection{Few-shot Learning and Meta-learning}

Our work is related to few-shot learning and meta-learning. The goal is to train deep learning models with diverse tasks, where each task is specified by a few examples for demonstration, so that the model can be quickly adapted to new tasks. One type of representative methods is the metric-based method~\cite{vinyals2016matching,snell2017prototypical,garcia2017few,sung2018learning}. The basic idea is to learn a prototype vector for each class based on the few examples, and use the prototype vectors for prediction. Another type of representative methods is the optimization-based method~\cite{finn2017model,ravi2016optimization}. Typically, these methods formalize the problem as a bi-level optimization problem~\cite{franceschi2018bilevel}. The outer loop learns global parameters shared across different tasks, such as the initialization of model parameters. The inner loop adapts the shared parameters to each task by performing several steps of gradient descent according to the few examples. Compared with these methods, which aim to learn a point estimation of prototype vectors or model parameters, our approach treats them as random variables and models their posterior distributions, which can thus handle the uncertainty of these prototype vectors or parameters.

In addition, there are several recent studies~\cite{kim2018bayesian,gordon2018meta,ravi2018amortized} also using Bayesian learning techniques for meta-learning, where the posterior distributions of prototype vectors or model parameters are inferred. However, these methods ignore the relationships of different classes, while we model their relationships by applying a graph neural network~\cite{kipf2016semi,gilmer2017neural,velivckovic2017graph} to a global graph of classes, allowing our approach to better generalize to all different classes. Furthermore, we model the posterior distribution in a more effective way. For \citet{gordon2018meta} and \citet{ravi2018amortized}, they use a simple Gaussian distribution parameterized by an amortization network to approximate the true posterior distribution. However, the true posterior distribution can be more complicated than the Gaussian distribution, and hence these methods are less precise. Another method from \citet{kim2018bayesian} uses Stein Variational Gradient Descent (SVGD)~\cite{liu2016stein} to draw samples from the posterior distribution for optimization, but SVGD relies on a properly-designed kernel function for different samples, which can be hard to choose. In contrast, our approach uses the stochastic gradient Langevin dynamics~\cite{welling2011bayesian} to perform Monte Carlo sampling for optimization, which is more flexible and effective as we will show in the experiment.

\subsection{Relation Extraction}

Relation extraction is a fundamental task in natural language processing. Given two entities in a sentence, the goal is to predict the relation expressed in the sentence. Most existing studies~\cite{zeng2014relation,zeng2015distant,zhang2017position} focus on the supervised or semi-supervised settings of relation extraction, where they assume massive labeled sentences are available. However, the number of labeled sentences can be very limited in practice. Some studies try to solve the challenge of insufficient labeled sentences by resorting to knowledge graphs, where existing facts are used to annotate unlabeled sentences through distant supervision~\cite{mintz2009distant} or provide additional training signals~\cite{shwartz2016improving,qu2018weakly,zhu2019integrating}. Nevertheless, the data or signals obtained in this way can be very noisy. Some more recent studies~\cite{han2018fewrel,gao2019fewrel,soares2019matching} instead focus on few-shot relation extraction, and the goal is to develop models which can be trained with only a few labeled sentences. By combining meta-learning methods with BERT encoders~\cite{devlin2018bert}, these methods achieve impressive results. However, they ignore the connections of different relations, which naturally exist in practice. In contrast, we treat a global graph of relations as prior knowledge, and propose a principled Bayesian meta-learning approach to leverage the global graph, which is able to better generalize across different relations.

\section{Problem Definition}

\begin{figure*}
	\centering
	\includegraphics[width=0.95\textwidth]{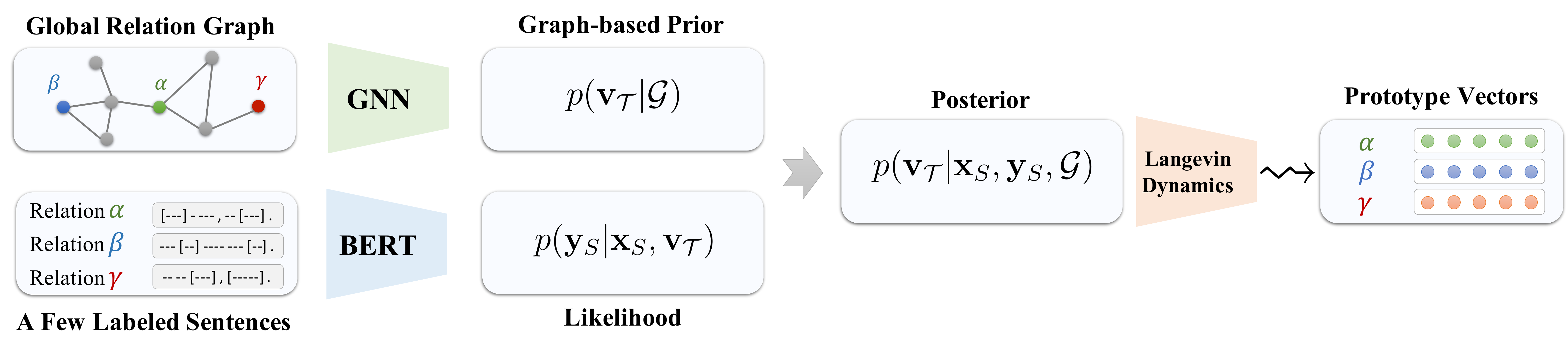}
	\caption{Framework overview. We consider a global relation graph and a few labeled sentences of each relation. Our approach aims at modeling the posterior distribution of prototype vectors for different relations. The prior distribution in the posterior is parameterized by applying a graph neural network to the global graph, and the likelihood is parameterized by using BERT to the labeled sentences. We use stochastic gradient Langevin dynamics to draw multiple samples from the posterior for optimization, which is in an end-to-end fashion.
	}
	\label{fig:framework}
\end{figure*}

Relation extraction is an important task in many research areas, which aims at predicting the relation of two entities given a sentence. Most existing methods for relation extraction require a large number of labeled sentences as training data, which are expensive to obtain. Therefore, more recent studies focus on few-shot relation extraction, where only a few examples for each relation are given as training data. However, the results are still far from satisfactory due to the limited information in these examples. To further improve the results, another data source should be considered.

In this paper, we propose to study few-shot relation extraction with a global graph of relations, where a global graph describing the relationships of all possible relations is assumed to be an additional data source. More formally, we denote the global relation graph as $\mathcal{G} = (\mathcal{R}, \mathcal{L})$, where $\mathcal{R}$ is the set of all the possible relations, and $\mathcal{L}$ is a collection of links between different relations. The linked relations are likely to have more similar semantic meanings. %The weight of each link measures the strength of the connection between the linked relations.

In few-shot relation extraction, each time we only consider a subset of relations from the whole relation set, i.e., $\mathcal{T} \subseteq  \mathcal{R}$. Given a few support sentences $\mathcal{S}$ of these relations, where $\mathbf{x}_{\mathcal{S}} = \{ \mathbf{x}_s\}_{s \in \mathcal{S}}$ represents the text of these sentences, and $\mathbf{y}_{\mathcal{S}} = \{ \mathbf{y}_s\}_{s \in \mathcal{S}}$ represents the corresponding labels with each $\mathbf{y}_s \in \mathcal{T}$, our goal is to learn a neural classifier for these relations by using the global graph and support sentences. Then given some unlabeled sentences as queries $\mathbf{x}_{\mathcal{Q}} = \{ \mathbf{x}_q\}_{q \in \mathcal{Q}}$, we will apply the classifier to predict their labels, i.e., $\mathbf{y}_{\mathcal{Q}} = \{ \mathbf{y}_q\}_{q \in \mathcal{Q}}$ with each $\mathbf{y}_q \in \mathcal{T}$.

\section{Model}

In this section, we introduce our approach for few-shot relation extraction with global relation graphs. Our approach represents each relation with a prototype vector, which can be used for classifying the query sentences. In contrast to most existing meta-learning methods, which learn a point estimation of the prototype vector, we treat the prototype vector as a random variable to model its posterior distribution. The posterior distribution is naturally composed of two terms, i.e., a prior of the prototype vector obtained from the global relation graph, and a likelihood function on the support sentences. Our approach parameterizes the prior distribution by applying a graph neural network to the global relation graph. By using such a graph-based prior, our approach can effectively generalize to different relations. However, optimization remains challenging in our approach, as the posterior distribution of prototype vectors has a complicated form. To address that, we approximate the posterior distribution through Monte Carlo sampling, where multiple samples of prototype vectors are drawn by using the stochastic gradient Langevin dynamics. By doing so, our approach can be effectively optimized in an end-to-end fashion.

\subsection{A Probabilistic Formalization}

Our approach uses Bayesian learning techniques for few-shot relation extraction, where we formalize the problem in a probabilistic way. More specifically, recall that given a subset of relations $\mathcal{T} \subseteq  \mathcal{R}$, the goal is to predict the labels $\mathbf{y}_Q$ of some query texts $\mathbf{x}_Q$ based on a global relation graph $\mathcal{G}$ and a few support sentences $(\mathbf{x}_S, \mathbf{y}_S)$. Formally, our goal can be stated as computing the following log-probability:
\begin{equation}
\label{eq:prob}
\begin{aligned}
    \log p(\mathbf{y}_Q | \mathbf{x}_Q, \mathbf{x}_S, \mathbf{y}_S, \mathcal{G}).
\end{aligned}
\end{equation}
We compute the probability by representing each relation $r\in\mathcal{T}$ with a prototype vector $\mathbf{v}_r$, which summarizes the semantic meaning of that relation. By introducing such prototype vectors, the log-probability can be factorized as:
\begin{equation}
\label{eq:prob-factorize}
\begin{aligned}
    &\log p(\mathbf{y}_Q | \mathbf{x}_Q, \mathbf{x}_S, \mathbf{y}_S, \mathcal{G}) \\
    =& \log \int p(\mathbf{y}_Q | \mathbf{x}_Q, \mathbf{v}_{\mathcal{T}}) p(\mathbf{v}_{\mathcal{T}}|\mathbf{x}_S, \mathbf{y}_S, \mathcal{G}) d \mathbf{v}_{\mathcal{T}}
\end{aligned}
\end{equation}
where $\mathbf{v}_{\mathcal{T}} = \{\mathbf{v}_r\}_{r \in \mathcal{T}}$ is a collection of prototype vectors for all the target relations in $\mathcal{T}$. These prototype vectors are characterized by both the labeled sentences in the support set and global relation graph through the distribution $p(\mathbf{v}_{\mathcal{T}}|\mathbf{x}_S, \mathbf{y}_S, \mathcal{G})$. With such prototype vectors to represent target relations, the distribution of query sentence labels can then be defined through a softmax function as follows:
\begin{equation}
\label{eq:pred}
\begin{aligned}
    &p(\mathbf{y}_Q | \mathbf{x}_Q, \mathbf{v}_{\mathcal{T}}) = \prod_{q \in Q} p(\mathbf{y}_q | \mathbf{x}_q, \mathbf{v}_{\mathcal{T}}),\ \text{with each}\\
    &p(\mathbf{y}_q = r | \mathbf{x}_q, \mathbf{v}_{\mathcal{T}}) = \frac{\exp(\mathcal{E}(\mathbf{x}_q) \cdot \mathbf{v}_r )}{\sum_{r' \in \mathcal{T}} \exp(\mathcal{E}(\mathbf{x}_q) \cdot \mathbf{v}_{r'})},
\end{aligned}
\end{equation}
where $\mathcal{E}$ is a sentence encoder, which encodes a query sentence $\mathbf{x}_q$ into an encoding $\mathcal{E}(\mathbf{x}_q)$. Intuitively, we compute the dot product of the encoding and the prototype vector $\mathbf{v}_r$ to estimate how likely the sentence expresses the relation.

Under such a formalization, the key is how to parameterize $p(\mathbf{v}_{\mathcal{T}}|\mathbf{x}_S, \mathbf{y}_S, \mathcal{G})$, which is the posterior distribution of prototype vectors conditioned on the support sentences and the global relation graph. Next, we introduce how we parameterize the posterior distribution in our proposed approach.

\subsection{Parameterization of the Posterior Distribution}

To model the posterior distribution of prototype vectors, we notice that the posterior can be naturally factorized into a prior distribution conditioned on the relation graph, and a likelihood function on the few support sentences. Therefore, we can formally represent the posterior as follows:
\begin{equation}
\label{eq:bayes}
\begin{aligned}
    p(\mathbf{v}_{\mathcal{T}}|\mathbf{x}_S, \mathbf{y}_S, \mathcal{G}) \propto p(\mathbf{y}_S | \mathbf{x}_S, \mathbf{v}_{\mathcal{T}}) p(\mathbf{v}_{\mathcal{T}} | \mathcal{G}),
\end{aligned}
\end{equation}
where $p(\mathbf{v}_{\mathcal{T}} | \mathcal{G})$ is the prior for the prototype vectors and $p(\mathbf{y}_S | \mathbf{x}_S, \mathbf{v}_{\mathcal{T}})$ is the likelihood on support sentences.

To extract the knowledge from the global relation graph to characterize the prior distribution, we introduce a graph neural network~\cite{kipf2016semi,gilmer2017neural,velivckovic2017graph,qu2019gmnn} in our approach. The graph neural network is denoted as $\mathcal{F}$, which can learn a latent representation $\mathbf{h}_r$ for each relation $r$, i.e., $\mathbf{h}_r = \mathcal{F}(\mathcal{G})_r$. More specifically, the graph neural network $\mathcal{F}$ initializes the latent embedding $\mathbf{h}_r$ of each relation as its initial feature vector. Then $\mathcal{F}$ iteratively updates the latent embedding of each relation $r$ according to the embeddings of $r$ and $r$'s neighbors. Formally, $\mathcal{F}$ updates the embeddings as follows:
\begin{equation}
\label{eq:gnn}
\begin{aligned}
    \mathbf{h}_r \leftarrow  \mathcal{U} \left\{ \sum_{r' \in \text{NB}(r)}\mathcal{M}(\mathbf{h}_{r'}), \mathbf{h}_r \right\},
\end{aligned}
\end{equation}
where $\text{NB}(r)$ is the neighbor of $r$ in the global graph, and $\mathcal{M}$ is a transformation function. Basically, for each relation $r$, we apply $\mathcal{M}$ to the latent embeddings of $r$'s neighbors and then aggregate the transformed embeddings together. Finally, the latent embedding of $r$ is updated based on its previous value and the aggregated embeddings through an update function $\mathcal{U}$. After several rounds of such update, the relationships between different relations encoded in the global graph can be effectively preserved into the final relation embeddings, which can serve as regularization for the prototype vectors. Motivated by that, we parameterize the prior distribution of prototype vectors $p(\mathbf{v}_{\mathcal{T}} | \mathcal{G})$ as follows:
\begin{equation}
\label{eq:prior}
\begin{aligned}
    p(\mathbf{v}_{\mathcal{T}} | \mathcal{G}) = \prod_{r \in \mathcal{T}} p(\mathbf{v}_r | \mathbf{h}_r) = \prod_{r \in \mathcal{T}} \mathcal{N}(\mathbf{v}_r | \mathbf{h}_r, \boldsymbol{I}),
\end{aligned}
\end{equation}
where we model the prior distribution of each relation $r \in \mathcal{T}$ independently. For each relation, we define its prior as a Gaussian distribution, where the mean is set as the latent representation $\mathbf{h}_r$ given by the graph neural network $\mathcal{F}$. In this way, the knowledge from the relation graph can be effectively distilled into the prior distribution, allowing our approach to better generalize to a wide range of relations.

Besides the graph-based prior, we also consider the likelihood on support sentences when parameterizing the posterior distribution of prototype vectors. Similar to the likelihood on the query sentences in Eq.~\eqref{eq:pred}, the likelihood on support sentences can be characterized as below:
\begin{equation}
\label{eq:likelihood}
\begin{aligned}
    &p(\mathbf{y}_S | \mathbf{x}_S, \mathbf{v}_{\mathcal{T}}) = \prod_{s \in S} p(\mathbf{y}_s | \mathbf{x}_s, \mathbf{v}_{\mathcal{T}}),\ \text{with each}\\
    &p(\mathbf{y}_s = r | \mathbf{x}_s, \mathbf{v}_{\mathcal{T}}) = \frac{\exp(\mathcal{E}(\mathbf{x}_s) \cdot \mathbf{v}_r )}{\sum_{r' \in \mathcal{T}} \exp(\mathcal{E}(\mathbf{x}_s) \cdot \mathbf{v}_{r'})},
\end{aligned}
\end{equation}
where $\mathcal{E}$ is the sentence encoder. By applying the likelihood on support sets to the prior distribution of prototype vectors, we can effectively adapt the prior distribution to the target relations with the few support sentences. In this way, the posterior distribution combines the knowledge from both the global relation graph and the support sentences, which can thus be used to effectively classify query sentences.

\subsection{Optimization and Prediction}
In the above section, we have introduced how we parameterize the posterior distribution of prototype vectors. Next, we explain the model optimization and prediction.

During both model optimization and prediction, we have to deal with the log-probability $\log p(\mathbf{y}_Q | \mathbf{x}_Q, \mathbf{x}_S, \mathbf{y}_S, \mathcal{G})$, where we either maximize or compute this value. However, according to Eq.~\eqref{eq:prob-factorize}, the log-probability relies on the integration over prototype vectors, which is intractable. Therefore, we estimate the log-probability with Monte Carlo sampling, where several samples of prototype vectors are drawn from the posterior distribution for approximation:
\begin{equation}
\label{eq:prob-monte-carlo}
\begin{aligned}
    &\log p(\mathbf{y}_Q | \mathbf{x}_Q, \mathbf{x}_S, \mathbf{y}_S, \mathcal{G}) \\
    =& \log \mathbb{E}_{p(\mathbf{v}_{\mathcal{T}}|\mathbf{x}_S, \mathbf{y}_S, \mathcal{G})}[p(\mathbf{y}_Q | \mathbf{x}_Q, \mathbf{v}_{\mathcal{T}})] \\
    \approx & \log \frac{1}{L}\sum_{l = 1}^L p(\mathbf{y}_Q | \mathbf{x}_Q, \mathbf{v}_{\mathcal{T}}^{(l)}).
\end{aligned}
\end{equation}
Here, $\mathbf{v}_{\mathcal{T}}^{(l)}$ is a sample drawn from the posterior distribution, i.e., $\mathbf{v}_{\mathcal{T}}^{(l)} \sim p(\mathbf{v}_{\mathcal{T}}|\mathbf{x}_S, \mathbf{y}_S, \mathcal{G})$. However, as shown in Eq.~\eqref{eq:bayes}, the posterior distribution combines both a graph-based prior and a likelihood function on support sentences. The graph-based prior is a Gaussian distribution while the likelihood function is specified by a softmax function. Therefore, the posterior distribution has a very complicated form, and sampling from the posterior is nontrivial. To address the problem, in this paper we use the stochastic gradient Langevin dynamics~\cite{welling2011bayesian}, which performs sampling via multiple stochastic updates. Formally, to draw a sample $\hat{\mathbf{v}}_{\mathcal{T}}$ of the prototype vector, we can randomly initialize the sample $\hat{\mathbf{v}}_{\mathcal{T}}$, and iteratively update the sample as follows:
\begin{equation}
\label{eq:sgmcmc-update}
\begin{aligned}
    \hat{\mathbf{v}}_{\mathcal{T}} \leftarrow  \hat{\mathbf{v}}_{\mathcal{T}} & + \frac{\epsilon}{2} \nabla_{\hat{\mathbf{v}}_{\mathcal{T}}}\log p(\mathbf{y}_S | \mathbf{x}_S, \hat{\mathbf{v}}_{\mathcal{T}}) p(\hat{\mathbf{v}}_{\mathcal{T}} | \mathcal{G})\\
    &+ \sqrt{\epsilon} \hat{\mathbf{z}},
\end{aligned}
\end{equation}
where $\hat{\mathbf{z}} \sim \mathcal{N}(\boldsymbol{0}, \boldsymbol{I})$ is a sample from the standard Gaussian distribution. After a burn-in period, $\hat{\mathbf{v}}_{\mathcal{T}}$ can then be treated as a true sample from the posterior distribution of prototype vectors. In the above update rule, the term $\nabla_{\hat{\mathbf{v}}_{\mathcal{T}}}\log p(\mathbf{y}_S | \mathbf{x}_S, \hat{\mathbf{v}}_{\mathcal{T}}) p(\hat{\mathbf{v}}_{\mathcal{T}} | \mathbf{h}_{\mathcal{T}})$ is highly related to the MAML algorithm~\cite{finn2017model}, as they both aim at moving the sample towards maximizing the likelihood $\log p(\mathbf{y}_S | \mathbf{x}_S, \hat{\mathbf{v}}_{\mathcal{T}})$ on support sentences, and thereby adapt to the target relations in $\mathcal{T}$. The difference is that we also leverage the graph-based prior $p(\hat{\mathbf{v}}_{\mathcal{T}} | \mathcal{G})$ to guide this process. Moreover, at each step, a random noise $\hat{\mathbf{z}}$ is added, allowing us to get different samples from the posterior distribution $p(\mathbf{v}_{\mathcal{T}}|\mathbf{x}_S, \mathbf{y}_S, \mathcal{G})$, rather than a single sample with maximum posterior probability. In other words, our approach is able to model the uncertainty of prototype vectors.

However, the Langevin dynamics requires a burn-in period, which can take a long time. To accelerate this process, it is helpful to let the samples stay in the high-density regions of the posterior distribution, such that we can better explore around those regions~\cite{gong2018meta}. Therefore, we try to initialize the sample $\hat{\mathbf{v}}_{\mathcal{T}}$ at a point with high posterior probability. Towards this goal, we theoretically justify in appendix that a proper initialization can be given as below:
\begin{equation}
\label{eq:sgmcmc-init}
\begin{aligned}
    \hat{\mathbf{v}}_{\mathcal{T}} &\leftarrow \{\hat{\mathbf{v}}_r\}_{r \in \mathcal{T}},\ \text{with each} \\
    \hat{\mathbf{v}}_r &\leftarrow  \mathbf{m}_r + \mathbf{h}_r - \mathbf{m},
\end{aligned}
\end{equation}
where $\mathbf{h}_r$ is the latent embedding of relation $r$ given by the graph neural network on the global relation graph, $\mathbf{m}_r$ is the mean encoding of all the sentences under relation $r$ in the support set, and $\mathbf{m}$ is the mean encoding of all the sentences in the support set. Intuitively, for each relation $r$, we add the latent embedding $\mathbf{h}_r$ from the global relation graph and the mean encoding $\mathbf{m}_r$ from the given examples of that relation. Also, we subtract the mean encoding $\mathbf{m}$ of all sentences in the support set, so that we can better distinguish sentences from different relations. In practice, we introduce two hyperparameters for $\mathbf{h}_r$ and $\mathbf{m}$ to control their relative weights. With such an initialization, we can empirically guarantee that the Langevin dynamics will quickly converge.

Once we obtain the samples of prototype vectors from their posterior distribution, we can then use the samples to optimize and compute $\log p(\mathbf{y}_Q | \mathbf{x}_Q, \mathbf{x}_S, \mathbf{y}_S, \mathcal{G})$ according to Eq.~\eqref{eq:prob-monte-carlo}. The whole optimization process is end-to-end, and we summarize the optimization algorithm in Alg.~\ref{alg::optim}.

\begin{algorithm}[bt]
    \caption{Training Algorithm}
    \label{alg::optim}
    \begin{algorithmic}
        \STATE {\bfseries Given:} A relation set $\mathcal{R}$, a global relation graph $\mathcal{G}$.
        \WHILE{not converge}
            \STATE 1. Sample a subset of relations $\mathcal{T} \subseteq \mathcal{R}$ as targets.
            \STATE 2. Sample support and query sets $(\mathbf{x}_{\mathcal{S}}, \mathbf{y}_{\mathcal{S}})\ (\mathbf{x}_{\mathcal{Q}}, \mathbf{y}_{\mathcal{Q}})$.
            \STATE 3. Compute the summary vectors $\mathbf{h}_{\mathcal{T}}$ of relations by applying the GNN $\mathcal{F}$ to the global relation graph $\mathcal{G}$.
            \STATE 4. Initialize prototype vectors $\{ \mathbf{v}_{\mathcal{T}}^{(l)} \}_{l=1}^L$ with Eq.~\eqref{eq:sgmcmc-init}.
            \STATE 5. Update prototype vectors for $M$ steps with Eq.~\eqref{eq:sgmcmc-update}.
            \STATE 6. Compute and maximize the log-likelihood function $\log p(\mathbf{y}_Q | \mathbf{x}_Q, \mathbf{x}_S, \mathbf{y}_S, \mathcal{G})$ based on Eq.~\eqref{eq:prob-monte-carlo}.
        \ENDWHILE
\end{algorithmic}
\end{algorithm}

\section{Experiment}

In this section, we empirically evaluate our proposed approach on two benchmark datasets, and we consider both the few-shot and the zero-shot learning settings.

\subsection{Datasets}

\begin{table}[bht]
    \vspace{-0.3cm}
	\caption{Dataset statistics.}
	\label{tab:dataset}
	\begin{center}
	\scalebox{0.85}
	{
		\begin{tabular}{C{4.0cm} C{2.0cm} C{2.0cm}}\hline
		\textbf{Dataset}	& \textbf{FewRel} & \textbf{NYT-20}  \\ 
		\hline
		\# All Possible Relations & 828 & 828 \\
		\# Training Relations & 64 & 10  \\
		\# Validation Relations & 16 & 5  \\
		\# Test Relations & 20 & 10  \\
		\# Sentences per Relation & 700 & 100  \\
		\hline
	    \end{tabular}
	}
	\vspace{-0.3cm}
	\end{center}
\end{table}

\begin{table}[bht]
	\caption{Results of few-shot classification on the FewRel test set (\%). We rerun all the algorithms with the same sentence encoder BERT$_{\small \texttt{BASE}}$. Our approach outperforms all the baseline methods.}
	\label{tab:result-wiki-test}
	\begin{center}
	\scalebox{0.8}
	{
		\begin{tabular}{C{2.3cm} | C{1.5cm} C{1.5cm} C{1.5cm} C{1.5cm}}\hline
		\textbf{Algorithm}	& \textbf{5-Way 1-Shot} & \textbf{5-Way 5-Shot} & \textbf{10-Way 1-Shot} & \textbf{10-Way 5-Shot} \\ 
		\hline
		GNN & 75.66 & 89.06 & 70.08 & 76.93 \\
		SNAIL & 67.75 & 86.58 & 54.29 & 77.54 \\
		Proto & 80.68 & 89.60 & 71.48 & 82.89 \\
		Pair & 88.32 & 93.22 & 80.63 & 87.02 \\
		MTB & 89.80 & 93.59 & 83.37 & 88.64 \\
		MAML & 89.70 & 93.55 & 83.17 & 88.51 \\
		Versa & 88.52 & 93.15 & 81.62 & 87.73 \\
		BMAML & 89.65 & 93.40 & 82.94 & 88.20 \\
		\hline
		REGRAB & \textbf{90.30} & \textbf{94.25} & \textbf{84.09} & \textbf{89.93} \\
		\hline
	    \end{tabular}
	}
	\vspace{-0.7cm}
	\end{center}
\end{table}

\begin{table*}[bht]
	\caption{Results of few-shot classification on FewRel validation set (\%). All the methods use the same encoder for fair comparison.}
	\label{tab:result-wiki-val}
	\begin{center}
	\scalebox{0.8}
	{
		\begin{tabular}{C{3.7cm}  C{3.0cm}  C{2.0cm} C{2.0cm} C{2.0cm} C{2.0cm}}\hline
		\textbf{Category} & \textbf{Algorithm}	& \textbf{5-Way 1-Shot} & \textbf{5-Way 5-Shot} & \textbf{10-Way 1-Shot} & \textbf{10-Way 5-Shot} \\ 
		\hline
		\multirow{4}{*}{\textbf{Meta-learning}} 
		& Pair & 85.66 & 89.48 & 76.84 & 81.76 \\
		& MTB & 84.61 & 88.76 & 75.22 & 80.15 \\
		& Proto & 82.92 & 91.32 & 73.24 & 83.68 \\
		& MAML & 82.93 & 86.21 & 73.20 & 76.06 \\
		\hline
		\multirow{3}{*}{\textbf{Bayesian Meta-learning}}
		&Versa & 84.47 & 88.44 & 74.70 & 79.20 \\
		&BMAML & 85.80 & 89.71 & 76.66 & 81.34 \\
		&REGRAB & \textbf{87.95} & \textbf{92.54} & \textbf{80.26} & \textbf{86.72} \\
		\hline
	    \end{tabular}
	}
	\end{center}
	\vspace{-0.3cm}
\end{table*}

\begin{table*}[bht]
	\caption{Results of few-shot classification on NYT-25 test set (\%). All the methods use the same sentence encoder for fair comparison.}
	\label{tab:result-nyt-test}
	\begin{center}
	\scalebox{0.8}
	{
		\begin{tabular}{C{3.7cm}  C{3.0cm}  C{2.0cm} C{2.0cm} C{2.0cm} C{2.0cm}}\hline
		\textbf{Category} & \textbf{Algorithm}	& \textbf{5-Way 1-Shot} & \textbf{5-Way 5-Shot} & \textbf{10-Way 1-Shot} & \textbf{10-Way 5-Shot} \\ 
		\hline
		\multirow{4}{*}{\textbf{Meta-learning}} 
		& Pair & 80.78 & 88.13 & 72.65 & 79.68 \\
		& MTB & 88.90 & 95.53 & 83.08 & 92.23 \\
		& Proto & 77.63 & 87.25 & 66.49 & 79.51 \\
		& MAML & 86.96 & 93.36 & 79.62 & 88.32 \\
		\hline
		\multirow{3}{*}{\textbf{Bayesian Meta-learning}}
		&Versa & 87.70 & 93.77 & 81.27 & 89.35 \\
		&BMAML & 87.03 & 93.90 & 79.74 & 88.72 \\
		&REGRAB & \textbf{89.76} & \textbf{95.66} & \textbf{84.11} & \textbf{92.48} \\
		\hline
	    \end{tabular}
	}
	\end{center}
	\vspace{-0.2cm}
\end{table*}

We use two benchmark datasets for evaluation. One dataset is the FewRel dataset~\cite{han2018fewrel,gao2019fewrel}, which has been recently proposed for few-shot relation extraction. Note that only the training set and validation set of FewRel are released, and the test set is not public, so researchers have to evaluate models on the remote cluster provided by the FewRel team. Because of that, we conduct most of the performance analysis on the validation set of FewRel, and for the test set we only report a final number. 

The other dataset is NYT-25. The raw data of NYT-25 is from the official website of FewRel~\footnote{~\url{https://github.com/thunlp/FewRel}}~\cite{han2018fewrel,gao2019fewrel}, where labeled sentences under 25 relations are provided by annotating the New York Times data. However, the dataset has not been splitted into training, validation and test sets. Therefore, we randomly sample 10 relations for training, 5 for validation and the remaining 10 for test.

For both datasets, the relations are from a knowledge graph named  Wikidata~\footnote{\url{https://www.wikidata.org/wiki/Wikidata:Main_Page}}, which has 828 unique relations in total. To construct the global graph of all the relations, we first employ GraphVite~\cite{zhu2019graphvite} to run the TransE algorithm~\cite{bordes2013translating} on Wikidata to learn a 512-dimensional embedding vector for each relation. Then we use the relation embeddings to construct a 10-nearest neighbor graph as the global relation graph, and the learned relation embeddings are treated as the initial relation features in GNN $\mathcal{F}$. The detailed statistics are summarized in Tab.~\ref{tab:dataset}.

\subsection{Compared Algorithms}

We choose the following few-shot relation extraction methods and meta-learning methods for comparison.

(1) \textbf{Pair}~\cite{gao2019fewrel}: A few-shot relation extraction method, which performs prediction by measuring the similarity of a pair of sentences. 
(2) \textbf{MTB}~\cite{soares2019matching}: A few-shot relation extraction method, which could be viewed as a variant of the prototypical network~\cite{snell2017prototypical}, where dot product is used to measure vector similarity rather than Euclidean distance. 
(3) \textbf{GNN}~\cite{garcia2017few}: A meta-learning method which uses graph neural networks for prediction. 
(4) \textbf{SNAIL}~\cite{mishra2017simple}: An algorithm which uses temporal convolutional neural networks and attention mechanisms for meta-learning. 
(5) \textbf{Proto}~\cite{snell2017prototypical}: The algorithm of prototypical networks. 
(6) \textbf{MAML}~\cite{finn2017model}: The model-agnostic meta-learning algorithm. 
(7) \textbf{Versa}~\cite{gordon2018meta}: A Bayesian meta-learning method which uses amortization networks to model the posterior of prototype vectors. 
(8) \textbf{BMAML}~\cite{kim2018bayesian}: A Bayesian meta-learning method which uses SVGD to learn the posterior distribution. 
(9) \textbf{REGRAB}: Our proposed approach for \underline{r}elation \underline{e}xtraction with \underline{gra}ph-based \underline{B}ayesian meta-learning.

For all the meta-learning algorithms, we use BERT$_{\small \texttt{BASE}}$~\cite{devlin2018bert} as encoder to project sentences into encodings, and apply a linear softmax classifier on top of the encoding for classification, where the meta-learning algorithms are used to learn the parameters in the classifier, or in other words the prototype vectors of different relations.

\subsection{Parameter Settings}

In our approach, we use BERT$_{\small \texttt{BASE}}$~\cite{devlin2018bert} as encoder to encode all the tokens in a sentence. Then we follow~\citet{soares2019matching} and combine the token encodings of entity mentions in a sentence as the sentence encoding. We do the same thing for the advanced meta-learning algorithms, i.e., MTB, Proto, MAML, Versa, BMAML, in order to make fair comparison. The details of how we compute sentence encodings are explained in the appendix. For the softmax function of the likelihood on support and query sentences, we apply an annealing temperature of 10. For the Gaussian prior of prototype vectors, we apply a one-layer graph convolutional network~\cite{kipf2016semi} to the global relation graph to compute the mean. We have also tried more layers, but only obtain very marginal improvement. For the stochastic gradient Langevin dynamics, the number of samples to draw is set as 10 by default, which is the same as used by other Bayesian meta-learning methods, and we perform 5 steps of update for these samples with the initial step size (i.e., $\epsilon$ in Eq.~\eqref{eq:sgmcmc-update}) as 0.1 by default. The graph encoder and sentence encoder are tuned by SGD with learning rate as 0.1. For other hyperparameters, they are selected on the FewRel validation set through grid search.

\subsection{Results}

\smallskip
\noindent \textbf{5.4.1. Comparison with Baseline Methods}

The main results on FewRel test set, FewRel validation set and NYT-25 test set are presented in Tab.~\ref{tab:result-wiki-test}, Tab.~\ref{tab:result-wiki-val} and Tab.~\ref{tab:result-nyt-test} respectively. For fair comparison, the same sentence encoder BERT$_{\small \texttt{BASE}}$ is used for all the compared approaches. From Tab.~\ref{tab:result-wiki-test}, we can see that the results of GNN and SNAIL are less competitive, showing that they are less effective to model textual data. Compared with Pair and MTB, which are specifically designed for few-shot relation extraction, our approach achieves relatively better results in all the tables, showing that our approach can better generalize to a variety of relations given a few examples. Besides, our approach also outperforms other meta-learning methods. Comparing our approach with MAML and the prototypical network (Proto), the performance gain mainly comes from two aspects. On the one hand, our approach considers a global graph of different relations, which provides prior knowledge about the relationships of all the relations, allowing our approach to better adapt to different relations. On the other hand, our approach uses a Bayesian learning framework, which effectively deals with the uncertainty of the prototype vectors for different relations. Moreover, our approach is also superior to other Bayesian meta-learning methods, i.e., Versa and BMAML. The reason is that we consider a graph-based prior in the posterior distribution, making our approach more powerful. Also, our approach performs optimization through Monte Carlo sampling with stochastic gradient Langevin dynamics, which models and optimizes the posterior distribution in a more effective way.

\smallskip
\noindent \textbf{5.4.2. Analysis of the Graph-based Prior}

Compared with existing studies on few-shot relation extraction, our approach uses a global relation graph. The relation graph provides knowledge of the relationship between different relations, allowing our approach to better generalize to various relations. To leverage such knowledge, our approach parameterizes the prior distribution of prototype vectors for different relations by applying a graph neural network to the relation graph. In this section, we conduct experiments to analyze the effect of such a graph-based prior distribution.

\begin{table}[bht]
	\caption{Analysis of the graph-based prior on FewRel validation set (\%). Removing the graph-based prior reduces the accuracy.}
	\label{tab:result-ablation-graph}
	\begin{center}
	\scalebox{0.8}
	{
		\begin{tabular}{C{4.8cm} C{2.0cm} C{2.0cm}}\hline
		\textbf{Algorithm}	& \textbf{5-Way 1-Shot} & \textbf{10-Way 1-Shot} \\ 
		\hline
		REGRAB with graph prior & \textbf{87.95} & \textbf{80.26} \\
		REGRAB w/o graph prior  & 85.82 & 77.70 \\
		\hline
	    \end{tabular}
	}
	\end{center}
\end{table}

We first conduct some ablation study on the FewRel validation set, where we compare two variants of our approach, i.e., with or without using the graph-based prior. The results are presented in Tab.~\ref{tab:result-ablation-graph}. We can see that removing the graph-based prior results in significantly worse results, showing the effect of the graph-based prior for improving the relation extraction performance in the few-shot learning setting.

\begin{figure}[tb!]
	\centering
	\scalebox{1.0}{
	\subfigure[FewRel Validation Set]{
		\label{fig::zero-wiki}
		\includegraphics[width=0.24\textwidth]{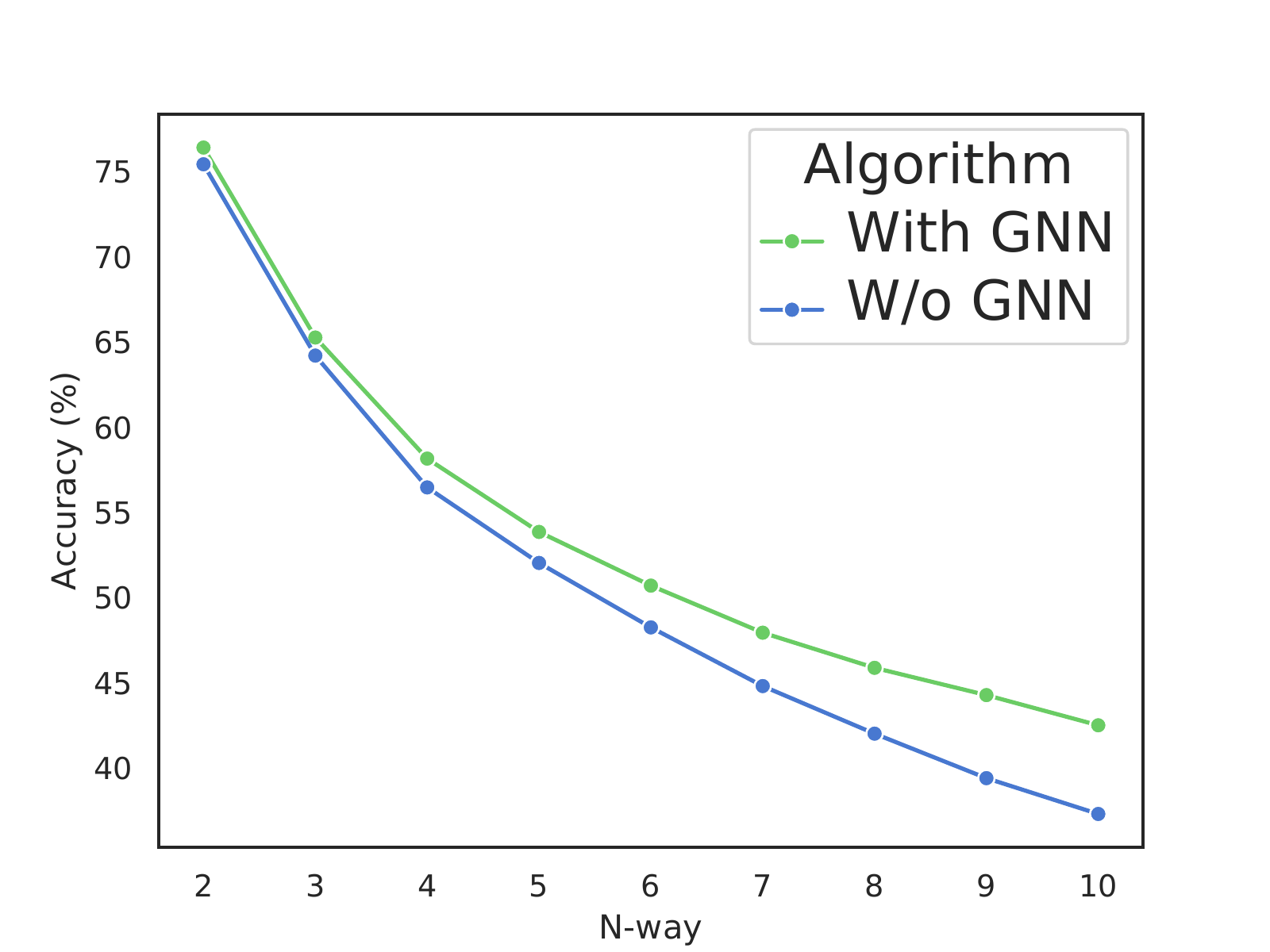}
	}
	\subfigure[NYT-25 Test Set]{
		\label{fig::zero-nyt}
		\includegraphics[width=0.24\textwidth]{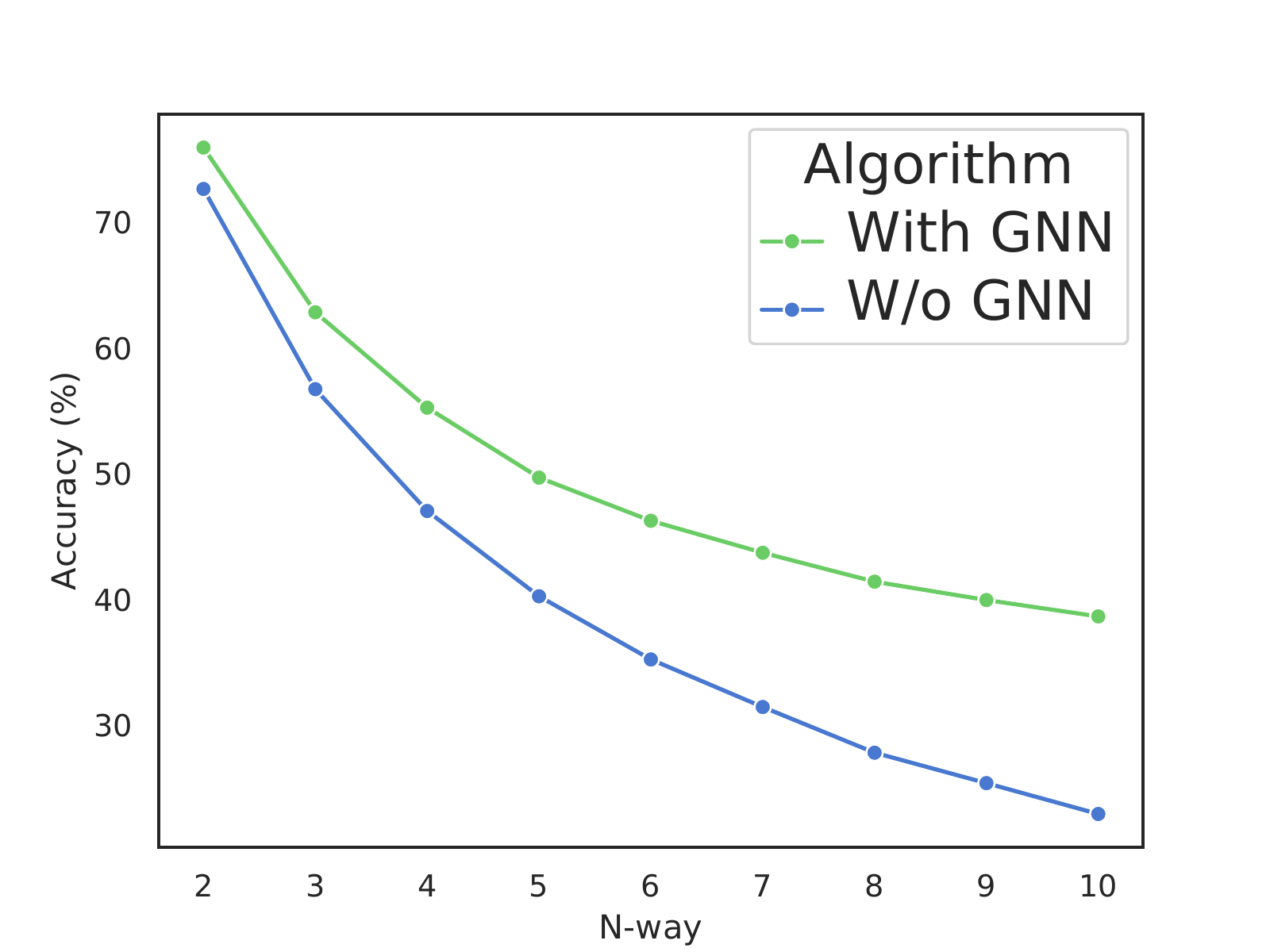}
	}
	}
	\caption{Results in zero-shot learning settings. Our approach can still achieve quite good results even without any labeled sentences.}
	\vspace{-0.5cm}
	\label{fig:zero}
\end{figure}

Moreover, with such a graph-based prior, our approach is able to handle relation extraction in the zero-shot learning setting, where no labeled sentences of each relation are given. Next, we present results on the FewRel validation set and NYT-25 test set to justify that. Remember our approach constructs the relation graph based on the pre-trained relation embeddings, and then applies a graph neural network for parameterizing the prior of prototype vectors. Compared with using graph neural networks, a more straightforward way is to directly apply a feedforward neural network to the pre-trained relation embeddings to derive the prior of prototype vectors. To demonstrate the advantange of graph neural networks, we also compare with the aforementioned variant. We present the results in Fig.~\ref{fig:zero}. From the figure, we can see that even without any labeled sentences as demonstration, our approach is still able to achieve quite impressive results, which proves its effectiveness. Also, comparing with the variant without using graph neural networks, our approach achieves significantly better results in both datasets. The observation shows that the graph neural network can help our approach better leverage the relationships of relations.

\smallskip
\noindent \textbf{5.4.3. Analysis of the Optimization Algorithm}

\begin{table}[bht]
    \vspace{-0.3cm}
	\caption{Ablation study of the optimization algorithm on FewRel validation set (\%). The Langevin dynamics leads to better results.}
	\label{tab:result-ablation-optimization}
	\begin{center}
	\scalebox{0.8}
	{
		\begin{tabular}{C{4.8cm} C{2.0cm} C{2.0cm}}\hline
		\textbf{Algorithm}	& \textbf{5-Way 1-Shot} & \textbf{10-Way 1-Shot} \\ 
		\hline
		REGRAB with Langevin & \textbf{87.95} & \textbf{80.26} \\
		REGRAB with Amortized VI & 85.74 & 77.41 \\
		\hline
	    \end{tabular}
	}
	%\vspace{-0.3cm}
	\end{center}
\end{table}

Compared with existing methods, the other difference of our approach is that we use the stochastic gradient Langevin dynamics during training, where several samples of prototype vectors are drawn for optimization. More specifically, we initialize a set of samples according to Eq.~\eqref{eq:sgmcmc-init}, and then perform multiple steps of update based on Eq.~\eqref{eq:sgmcmc-update}. In this part, we thoroughly analyze the optimization algorithm.

We start with the ablation study on the FewRel validation set, where we compare with a variant which parameterizes the posterior of prototype vectors as a Gaussian distribution. The mean of the Gaussian distribution is set as the value given by Eq.~\eqref{eq:sgmcmc-init}, which is the same as the initialization of samples we use in the Langevin dynamics. Such a variant is similar to the amortized variational inference method used in Versa~\cite{gordon2018meta}, an existing Bayesian meta-learning algorithm. We present the results in Tab.~\ref{tab:result-ablation-optimization}. We see that our approach with Langevin dynamics achieves relatively better results than the variant with amortized variational inference, which proves the effectiveness of approximating the posterior distribution of prototype vectors by drawing samples with the Langevin dynamics.

\begin{figure}[tb!]
	\centering
	\scalebox{1.0}{
	\subfigure[Analysis of \#Samples]{
		\label{fig::param-smp}
		\includegraphics[width=0.24\textwidth]{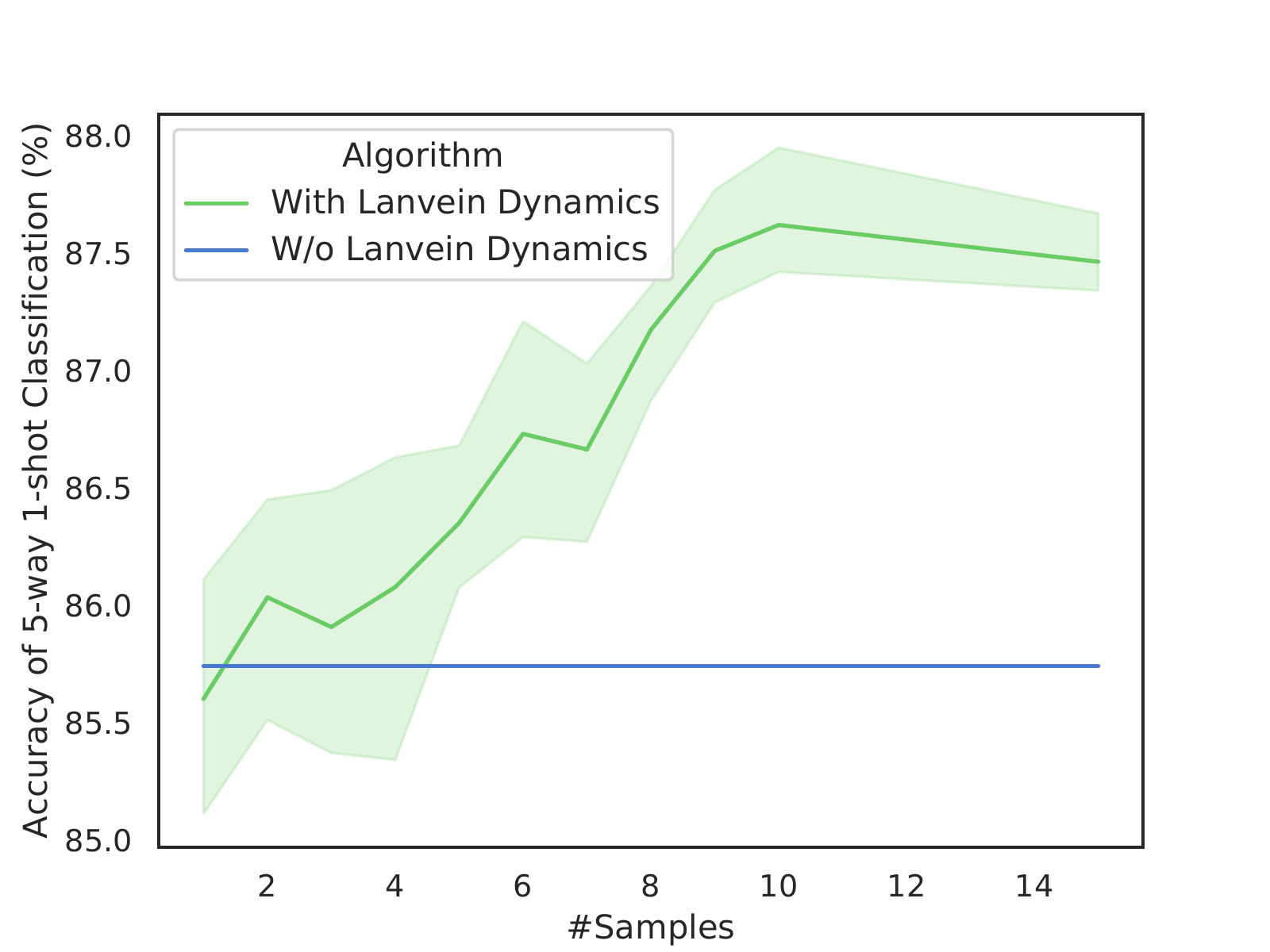}
	}
	\subfigure[Analysis of \#Steps]{
		\label{fig::param-step}
		\includegraphics[width=0.24\textwidth]{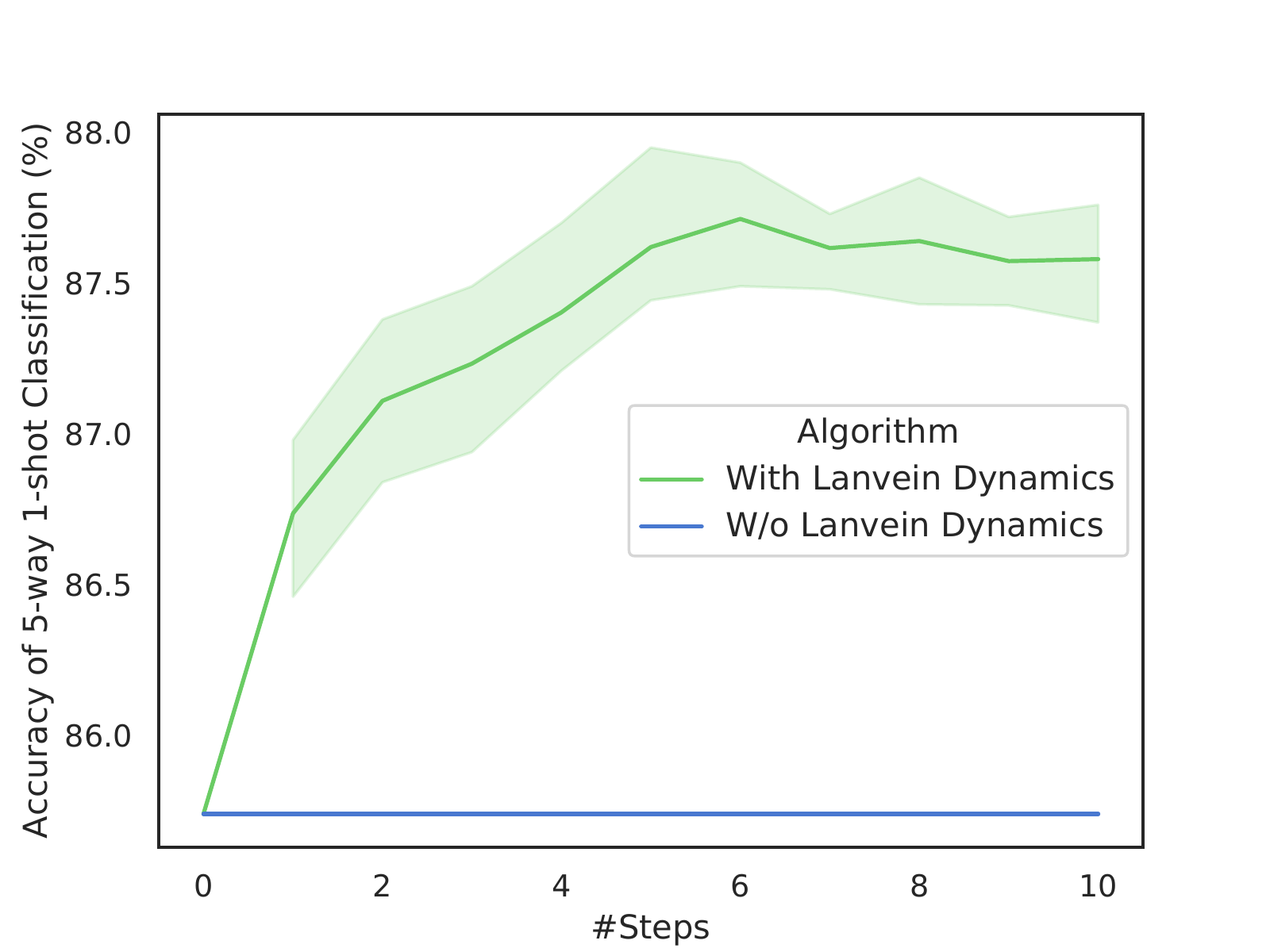}
	}
	}
	\caption{Performance w.r.t. the number of samples and the number of steps in the stochastic gradient Langevin dynamics.}
	\vspace{-0.5cm}
	\label{fig:param}
\end{figure}

In addition, when drawing samples from the posterior distribution, the Langevin dynamics performs multiple steps of update on a set of samples. Therefore two important hyperparameters of this process are the number of samples and the number of update steps. Next, we analyze these hyperparameters by doing sensitivity analysis. We take the FewRel validation set as an example, and report the accuracy of 5-way 1-shot classification. To better understand the result, we introduce a variant, where we only initialize the samples of prototype vectors through Eq.~\eqref{eq:sgmcmc-init}, without further updating them according to Eq.~\eqref{eq:sgmcmc-update}. Fig.~\ref{fig::param-smp} shows the results under different numbers of samples. We can see if only one or two samples are used, the results are quite poor, which are even worse than the variant without updating the samples. The reason is that if we only use few samples, the estimation of the log-probability in Eq.~\eqref{eq:prob-monte-carlo} can have high variance, leading to poor results. As we use more samples, the results are quickly improved, and the results converge when around 8-10 samples are used, which is quite efficient. Besides, Fig.~\ref{fig::param-step} presents the results under different numbers of update steps. As the number of steps is increased, the accuracy also rises, since the samples are moving towards high-density regions of the posterior to perform exploration. After only 4-5 steps, the accuracy quickly converges, which is very efficient. This observation proves the effectiveness of our initialization strategy presented in the Eq.~\eqref{eq:sgmcmc-init}.

\section{Conclusion}

This paper studies relation extraction in the few-shot learning setting, and the key idea is to consider a global relation graph, which captures the global relationship between relations. We propose a novel Bayesian meta-learning approach, which aims to model the posterior distribution of prototype vectors for different relations. The prior distribution in the posterior is parameterized by applying a graph neural network to the global relation graph. The stochastic gradient Langevin dynamics is used to optimize the posterior distribution. Experiments on two datasets prove the effectiveness of our approach. In the future, we plan to study learning the structure of the relation graph automatically by following existing studies~\cite{franceschi2019learning}. Besides, we also plan to apply our approach to other applications, such as few-shot image classification~\cite{triantafillou2019meta}.

\section*{Acknowledgements}

This project is supported by the Natural Sciences and Engineering Research Council (NSERC) Discovery Grant, the Canada CIFAR AI Chair Program, collaboration grants between Microsoft Research and Mila, Amazon Faculty Research Award, Tencent AI Lab Rhino-Bird Gift Fund and a NRC Collaborative R\&D Project (AI4D-CORE-06). We would like to thank Andreea Deac and Zhaocheng Zhu for reviewing the paper before submission.

\nocite{simic2008global}

\bibliography{paper}
\bibliographystyle{icml2020}

%\newpage

\appendix

\onecolumn

\section{Justification of the Initialization in the Langevin Dynamics}

In our approach, we leverage the Langevin dynamics to draw samples of the prototype vectors from their posterior distributions. Recall that the Langevin dynamics requires a burn-in period, which can take a long time. To accelerate the process, it is helpful to initialize the sample within a high-density area, which can prevent the sample from exploring low-density areas. Motivated by that, our approach aims to initialize the sample of prototype vector at a point with high posterior probability, and therefore we use the following strategy for initialization:
\begin{equation}\nonumber
\label{eq:appendix-sgmcmc-init}
\begin{aligned}
    \hat{\mathbf{v}}_{\mathcal{T}} &\leftarrow \{\hat{\mathbf{v}}_r\}_{r \in \mathcal{T}},\ \text{with each} \\
    \hat{\mathbf{v}}_r &\leftarrow  \mathbf{m}_r + \mathbf{h}_r - \mathbf{m},
\end{aligned}
\end{equation}
where $\mathbf{m}$ is the mean encoding of all the sentences in the support set, and $\mathbf{m}_r$ is the mean encoding for all the sentences of relation $r$ in the support set. In the remainder of this section we justify this choice.

Our goal is to find the point with high posterior probability. Suppose we consider the $N$-way $K$-shot setting, where there are $N$ relations in $\mathcal{T}$ (i.e., $|\mathcal{T}|=N$), and each relation has $K$ examples in the support set. Then the posterior is given as:
\begin{equation}\nonumber
\label{eq:appendix-log-posterior}
\begin{aligned}
    &\log p(\mathbf{v}_{\mathcal{T}}|\mathbf{x}_S, \mathbf{y}_S, \mathcal{G}) \\
    =& \frac{1}{K} \log p(\mathbf{y}_S | \mathbf{x}_S, \mathbf{v}_{\mathcal{T}}) + \log p(\mathbf{v}_{\mathcal{T}} | \mathcal{G}) + \text{const} \\
    =& \frac{1}{K} \sum_{s \in \mathcal{S}} \sum_{r \in \mathcal{T}} \mathbb{I}\{ \mathbf{y}_s = r \} \log \frac{\exp(\mathcal{E}(\mathbf{x}_s) \cdot \mathbf{v}_r)}{\sum_{r' \in \mathcal{T}} \exp(\mathcal{E}(\mathbf{x}_s) \cdot \mathbf{v}_{r'})}+\sum_{r \in \mathcal{T}}\log \exp\left(-\frac{1}{2}\left \| \mathbf{v}_r - \mathbf{h}_r \right \|_2^2\right) + \text{const} \\
    =& \frac{1}{K}\sum_{s \in \mathcal{S}} \sum_{r \in \mathcal{T}} \mathbb{I}\{ \mathbf{y}_s = r \} (\mathcal{E}(\mathbf{x}_s) \cdot \mathbf{v}_r) - \frac{1}{K}\sum_{s \in \mathcal{S}} \log \sum_{r \in \mathcal{T}} \exp(\mathcal{E}(\mathbf{x}_s) \cdot \mathbf{v}_r) - \frac{1}{2} \sum_{r \in \mathcal{T}} \left \| \mathbf{v}_r - \mathbf{h}_r \right \|_2^2 + \text{const},
\end{aligned}
\end{equation}
where we add a normalization term $\frac{1}{K}$ to the log-likelihood function, which makes the log-likelihood numerically stable as we increase the number of examples for each relation (i.e., $K$).

Our goal is to find a point of $\mathbf{v}_{\mathcal{T}}$ to maximize the above log-probability. However, the log-probability contains a log-partition function $\log \sum_{r \in \mathcal{T}} \exp(\mathcal{E}(\mathbf{x}_s) \cdot \mathbf{v}_r)$, which is hard to compute. To address this challenge, we aim at deriving a lower bound of the log-probability function for approximation. For this purpose, in this paper we use an inequation proposed by~\citet{simic2008global}, which is formally stated in the following theorem.
\begin{theorem}
Suppose that $\tilde{x} = \{ x_i \}_{i=1}^n$ represents a finite sequence of real numbers belonging to a fixed closed interval $I = [a,b]$, $a < b$. If $f$ is a convex function on $I$, then we have that:
\begin{equation}\nonumber
\frac{1}{n}\sum_{i=1}^n f(x_i) - f\left(\frac{1}{n}\sum_{i=1}^n x_i\right) \leq f(a) + f(b) - 2f\left(\frac{a + b}{2}\right).
\end{equation}
\end{theorem}
Based on the above theorem, we can have the following corollary:
\begin{corollary}
Suppose that for all the $s \in \mathcal{S}$ and $r \in \mathcal{T}$, we have $\exp(\mathcal{E}(\mathbf{x}_s) \cdot \mathbf{v}_r) \in [a, b]$. As $(-\log)$ is a convex function, we therefore have:
\begin{equation}\nonumber
- \frac{1}{|\mathcal{T}|} \sum_{r \in \mathcal{T}} \log(\exp(\mathcal{E}(\mathbf{x}_s) \cdot \mathbf{v}_r)) + \log\left(\frac{1}{|\mathcal{T}|}\sum_{r \in \mathcal{T}}  \exp(\mathcal{E}(\mathbf{x}_s) \cdot \mathbf{v}_r)\right) \leq -\log(a) -\log(b) + 2\log\left(\frac{a + b}{2}\right).
\end{equation}
After some simplification, we get:
\begin{equation}\nonumber
\begin{aligned}
\log\left(\sum_{r \in \mathcal{T}} \exp(\mathcal{E}(\mathbf{x}_s) \cdot \mathbf{v}_r)\right) \leq& \sum_{r \in \mathcal{T}} \frac{1}{|\mathcal{T}|} \log(\exp(\mathcal{E}(\mathbf{x}_s) \cdot \mathbf{v}_r)) + \log(|\mathcal{T}|) -\log(a) -\log(b) + 2\log\left(\frac{a + b}{2}\right)\\
=& \sum_{r \in \mathcal{T}} \frac{1}{|\mathcal{T}|}\mathcal{E}(\mathbf{x}_s) \cdot \mathbf{v}_r + \log(|\mathcal{T}|) -\log(a) -\log(b) + 2\log\left(\frac{a + b}{2}\right).
\end{aligned}
\end{equation}
\end{corollary}
In practice, we can easily find such $a$ and $b$ so that $\exp(\mathcal{E}(\mathbf{x}_s) \cdot \mathbf{v}_r) \in [a, b]$ is satisfied. Given this corollary, we are able to obtain a lower bound of the log-proability function as follows:
\begin{equation}\nonumber
\label{eq:appendix-lower-bound}
\begin{aligned}
    &\log p(\mathbf{v}_{\mathcal{T}}|\mathbf{x}_S, \mathbf{y}_S, \mathcal{G}) \\
    =& \frac{1}{K} \sum_{s \in \mathcal{S}} \sum_{r \in \mathcal{T}} \mathbb{I}\{ \mathbf{y}_s = r \} (\mathcal{E}(\mathbf{x}_s) \cdot \mathbf{v}_r) - \frac{1}{K} \sum_{s \in \mathcal{S}} \log \sum_{r \in \mathcal{T}} \exp(\mathcal{E}(\mathbf{x}_s) \cdot \mathbf{v}_r) - \frac{1}{2} \sum_{r \in \mathcal{T}} \left \| \mathbf{v}_r - \mathbf{h}_r \right \|_2^2 +\text{const} \\
    \geq & \frac{1}{K} \sum_{s \in \mathcal{S}} \sum_{r \in \mathcal{T}} \mathbb{I}\{ \mathbf{y}_s = r \} (\mathcal{E}(\mathbf{x}_s) \cdot \mathbf{v}_r) - \frac{1}{K} \sum_{s \in \mathcal{S}} \left[ \sum_{r \in \mathcal{T}} \frac{1}{|\mathcal{T}|} \mathcal{E}(\mathbf{x}_s) \cdot \mathbf{v}_r + \log(|\mathcal{T}|) - \log(a) -\log(b) + 2\log\left(\frac{a + b}{2}\right) \right] \\
    &- \frac{1}{2} \sum_{r \in \mathcal{T}} \left \| \mathbf{v}_r - \mathbf{h}_r \right \|_2^2+ \text{const} \\
    =& \frac{1}{K} \sum_{s \in \mathcal{S}} \sum_{r \in \mathcal{T}} \mathbb{I}\{ \mathbf{y}_s = r \} (\mathcal{E}(\mathbf{x}_s) \cdot \mathbf{v}_r) - \frac{1}{K} \sum_{s \in \mathcal{S}} \sum_{r \in \mathcal{T}} \frac{1}{N} \mathcal{E}(\mathbf{x}_s) \cdot \mathbf{v}_r - \frac{1}{2} \sum_{r \in \mathcal{T}} \left \| \mathbf{v}_r - \mathbf{h}_r \right \|_2^2 + \text{const}.
\end{aligned}
\end{equation}
Based on that, let us denote $\mathbf{m} = \frac{1}{NK} \sum_{s \in \mathcal{S}} \mathcal{E}(\mathbf{x}_s)$ to be the mean of encodings for all the sentences in the support set, and denote $\mathbf{m}_r = \frac{1}{K} \sum_{s \in \mathcal{S}} \mathbb{I}\{ \mathbf{y}_s = r \}\mathcal{E}(\mathbf{x}_s)$ to be the mean of encodings for sentences under relation $r$ in the support set. In this way, the above lower bound can be rewritten as follows:
\begin{equation}\nonumber
\label{eq:appendix-lower-bound-2}
\begin{aligned}
&\log p(\mathbf{v}_{\mathcal{T}}|\mathbf{x}_S, \mathbf{y}_S, \mathcal{G}) \\
\geq& \frac{1}{K} \sum_{s \in \mathcal{S}} \sum_{r \in \mathcal{T}} \mathbb{I}\{ \mathbf{y}_s = r \} (\mathcal{E}(\mathbf{x}_s) \cdot \mathbf{v}_r) - \frac{1}{K} \sum_{s \in \mathcal{S}} \sum_{r \in \mathcal{T}} \frac{1}{N}\mathcal{E}(\mathbf{x}_s) \cdot \mathbf{v}_r - \frac{1}{2} \sum_{r \in \mathcal{T}} \left \| \mathbf{v}_r - \mathbf{h}_r \right \|_2^2 + \text{const} \\
=& \sum_{r \in \mathcal{T}} \left[ \mathbf{v}_r \cdot \mathbf{m}_r - \mathbf{v}_r \cdot \mathbf{m} +  \mathbf{v}_r \cdot \mathbf{h}_r - \frac{1}{2} \mathbf{v}_r \cdot \mathbf{v}_r - \frac{1}{2} \mathbf{h}_r \cdot \mathbf{h}_r  \right] + \text{const} \\
=& \sum_{r \in \mathcal{T}} \left[ \mathbf{v}_r \cdot \mathbf{m}_r - \mathbf{v}_r \cdot \mathbf{m} + \mathbf{v}_r \cdot \mathbf{h}_r - \frac{1}{2} \mathbf{v}_r \cdot \mathbf{v}_r   \right] + \text{const} \\
=& \sum_{r \in \mathcal{T}} \left[  -\frac{1}{2} \left \| \mathbf{v}_r -  \mathbf{m}_r - \mathbf{h}_r +  \mathbf{m} \right \|_2^2 \right] + \text{const}.
\end{aligned}
\end{equation}
Therefore, under this lower bound, the optimal initialization of prototype vector $\mathbf{v}_r$ for each relation $r \in \mathcal{T}$ is given by:
\begin{equation}\nonumber
\label{eq:appendix-init}
\mathbf{v}_r^* = \mathbf{m}_r + \mathbf{h}_r - \mathbf{m},
\end{equation}
where we can ensure $\mathbf{v}_r^*$ to have a pretty high probability under the posterior distribution, and hence the Langevin dynamics is likely to converge faster.

\section{Details on the Sentence Encoder}

In our approach, we use the entity marker method proposed in \citet{soares2019matching} to generate the encoding of each sentence with BERT$_{\small \texttt{BASE}}$~\cite{devlin2018bert}. More specifically, recall that the goal of relation extraction is to predict the relation between two entities expressed in a sentence. Therefore, each sentence contains two entity mentions, i.e., token spans corresponding to an entity. For example, a sentence can be ``\textbf{Washington} is the capital of the \textbf{United States} .'', where \textbf{Washington} and \textbf{United States} are the entity mentions. During preprocessing, we follow \citet{soares2019matching} and add two markers for each entity mention, including a starting marker before the entity mention and an ending marker after the entity mention. In this way, the example sentence becomes ``\emph{[E1]} \textbf{Washington} \emph{[/E1]} is the capital of the \emph{[E2]} \textbf{United States} \emph{[/E2]} .''. Here, \emph{[E1]} and \emph{[E2]} are the starting markers. \emph{[/E1]} and \emph{[/E2]} are the ending markers. Then we apply BERT$_{\small \texttt{BASE}}$ to the preprocessed sentence, yielding an embedding vector for each token in the sentence. Finally, we follow \citet{soares2019matching} to concatenate the embeddings of token \emph{[E1]} and token \emph{[E2]} as the sentence encoding.

\section{Comparison of Similarity Measures}

In our approach, given the encoding $\mathcal{E}(\mathbf{x})$ of a sentence $\mathbf{x}$ and relation prototype vectors $\mathbf{v}_{\mathcal{T}}$, we predict the label $\mathbf{y}$ as below:
\begin{equation}\nonumber
    p(\mathbf{y} = r | \mathbf{x}, \mathbf{v}_{\mathcal{T}}) = \frac{\exp(\mathcal{E}(\mathbf{x}) \cdot \mathbf{v}_r )}{\sum_{r' \in \mathcal{T}} \exp(\mathcal{E}(\mathbf{x}) \cdot \mathbf{v}_{r'})},
\end{equation}
where we compute the dot product between sentence encodings and relation prototype vectors, and treat the value as logits for classification. Intuitively, the dot product could be understood as a similarity measure between encodings and prototype vectors. Besides dot product, Euclidean distance is another widely-used similarity measure, and we could naturally change the similarity measure in our approach to Euclidean distance as follows:
\begin{equation}\nonumber
    p(\mathbf{y} = r | \mathbf{x}, \mathbf{v}_{\mathcal{T}}) = \frac{\exp(-\frac{1}{2} \left \| \mathcal{E}(\mathbf{x}) - \mathbf{v}_r ) \right \|^2 )}{\sum_{r' \in \mathcal{T}} \exp(-\frac{1}{2} \left \| \mathcal{E}(\mathbf{x}) - \mathbf{v}_{r'} \right \|^2)}.
\end{equation}
In this section, we empirically compare the results of the two similarity measures, where the same configuration of hyperparameters is used for both similarity measures. The results are presented in Tab.~\ref{tab::similarity1} and Tab.~\ref{tab::similarity2}, where we can see that dot product works better in the 1-shot learning setting, whereas Euclidean distance achieves higher accuracy in the 5-shot learning setting. Therefore, when the number of support sentences for each relation is very limited (e.g., 1 or 2), it is better to use dot product. When we have more support sentences (e.g., 5 or more per relation), Euclidean distance is a better choice.

\begin{center}
    \begin{minipage}{0.48\linewidth}
        \centering
        \captionof{table}{Results on FewRel test set.}
        \scalebox{0.7}
	    {
		    \begin{tabular}{C{3cm} | C{1.5cm} C{1.5cm} C{1.5cm} C{1.5cm}}\hline
		    \textbf{Similarity Measure}	& \textbf{5-Way 1-Shot} & \textbf{5-Way 5-Shot} & \textbf{10-Way 1-Shot} & \textbf{10-Way 5-Shot} \\ 
		    \hline
		    Dot Product & \textbf{90.30} & 94.25 & \textbf{84.09} & \textbf{89.93} \\
		    Euclidean Distance & 86.74 & \textbf{94.34} & 78.56 & 88.95 \\
		    \hline
	        \end{tabular}
	    }
        \label{tab::similarity1}
    \end{minipage}
    \hfill
    \begin{minipage}{0.48\linewidth}
        \centering
        \captionof{table}{Results on FewRel validation set.}
        \scalebox{0.7}
	    {
		    \begin{tabular}{C{3cm} | C{1.5cm} C{1.5cm} C{1.5cm} C{1.5cm}}\hline
		    \textbf{Similarity Measure}	& \textbf{5-Way 1-Shot} & \textbf{5-Way 5-Shot} & \textbf{10-Way 1-Shot} & \textbf{10-Way 5-Shot} \\ 
		    \hline
		    Dot Product & \textbf{87.95} & 92.54 & \textbf{80.26} & 86.72 \\
		    Euclidean Distance & 86.79 & \textbf{94.44} & 78.48 & \textbf{88.92} \\
		    \hline
	        \end{tabular}
	    }
        \label{tab::similarity2}
    \end{minipage}
\end{center}

\end{document}